\documentclass[conference]{IEEEtran}
 \IEEEoverridecommandlockouts
\usepackage{cite}
\usepackage{amsmath,amssymb,amsfonts}
\usepackage[ruled,vlined,linesnumbered,noresetcount ]{algorithm2e}
\usepackage{graphicx}
\usepackage{textcomp}
\usepackage{xcolor}
\def\BibTeX{{\rm B\kern-.05em{\sc i\kern-.025em b}\kern-.08em
    T\kern-.1667em\lower.7ex\hbox{E}\kern-.125emX}}
  
 \usepackage{microtype}  
\usepackage{mathtools}  
\usepackage{tabulary}
  \usepackage{lipsum}
\usepackage{ nicefrac}
\usepackage{bbm} 
\usepackage{comment} 
\usepackage{xcolor}

\def\red{\color{red}}

\usepackage[hidelinks]{hyperref}
\definecolor{darkred}{RGB}{150,0,0}
\definecolor{darkgreen}{RGB}{0,150,0}
\definecolor{darkblue}{RGB}{0,0,150}
\hypersetup{colorlinks=true, linkcolor=red, citecolor=blue, urlcolor=darkblue}

\newtheorem{assumption}{Assumption}
\newtheorem{theorem}{Theorem}[section]

\newtheorem{Proposition}[theorem]{Proposition}

\newcommand{\norm}[1]{\left\lVert#1\right\rVert}

\begin{document}

\title{Multi-Environment Meta-Learning in Stochastic Linear  Bandits}

\author{Ahmadreza Moradipari$^1$, Mohammad Ghavamzadeh$^2$, Taha Rajabzadeh$^3$, \\ Christos Thrampoulidis$^4$, Mahnoosh Alizadeh$^1$\\


\thanks{This work is supported by NSF grant 1847096. C. Thrampoulidis was partially supported by the
NSF under Grant Number 1934641.
$^{1}$University of California, Santa Barbara,
$^{2}$Google Research,
%
$^{3}$Stanford University,
$^{4}$University of British Columbia.
Corresponding author: {\tt\small ahmadreza$\_$moradipari@ucsb.edu}}
}




\maketitle
\begin{abstract}
In this work we investigate meta-learning (or learning-to-learn) approaches in multi-task linear stochastic bandit problems that can originate from multiple environments. Inspired by the work of \cite{cella2020meta} on meta-learning in a sequence of linear bandit problems whose parameters are sampled from a single  distribution (i.e., a single environment), here we consider the feasibility of meta-learning when task parameters are drawn from a mixture distribution instead. For this problem, we propose a regularized version of the OFUL algorithm that, when trained on tasks with labeled environments, achieves low regret on a new task without requiring knowledge of the environment from which the new task originates. Specifically, our regret bound for the new algorithm  captures the effect of environment misclassification and highlights the benefits over  learning each task separately or meta-learning without recognition of the distinct mixture components. 
\end{abstract}



\section{Introduction}
Stochastic bandit optimization algorithms have long found applications in many fields where some characteristics of the users' response are not known and can only be learnt through a limited number of  noisy observations, including recommendation engines, advertisement placement, personalized medicine, etc. The learner's objective for the overall learning task consists of maximizing the cumulative reward gained during $T$ rounds of interaction with the user. The expected reward gained at each round $t$ is a function $f(x_t)$ of the action $x_t$ that the learner chooses to play, and $f$ is not known to the learner.  There is a rich literature covering parametric or non-parametric characterizations of $f$, as well as finite or continuous action sets. An important and widely applicable case is the stochastic linear bandit (LB) problem, where the expected reward is linear in the action $x_t$, i.e., $f(x_t) = x_t^T\theta$, with   $\theta$ denoting an unknown vector that describes the users' characteristics.

Now consider the scenario where a recommendation system consecutively deals with users whose characteristics (e.g., the parameter vector  $\theta$ in the LB case) originate from an \textit{unknown} probability distribution $\rho$. A classical bandit algorithm would approach each learning task independently, which would translate to high exploration cost to estimate $\theta$ for each user. Motivated by this, the  work of \cite{cella2020meta} explores the idea of transferring knowledge between consecutive tasks by designing a meta-learning algorithm for the LB problem. Meta-learning approaches, recently made popular in the reinforcement learning literature in order to address sample complexity issues, allow algorithms to acquire inductive biases in a data-driven manner in order to adapt faster to new situations based on their limited past experience.

In this work we consider the case where the user population's preferences originate from a mixture model, with sub-populations that have distinct (unknown) preference distributions. In this case, we say that the consecutive learning tasks originate from multiple environments.  Consider for example a recommendation system where men and women may have distinct preferences (e.g. on Netflix, some movies are seen to be preferred by more women and others by more men). Focused on the LB problem, we show that if the sub-populations' preference distributions are sufficiently distinct, in order to best transfer knowledge across learning tasks, the meta-learning algorithm should first estimate the environment from which the task originates. Our proposed algorithm MEML-OFUL, and the corresponding regret guarantees, formalize the trade-offs associated with this design choice.

Before formally stating the problem, let us provide an overview of prior art under three relevant categories.\\
\textbf{Multi-armed Bandits (MAB).~~~}
Two popular algorithms exist for MAB: 1)    the   upper confidence bound (UCB) algorithm based on the optimism in the face of uncertainty (OFU) principle \cite{Auer,li2010contextual,filippi2010parametric}, which chooses the best feasible environment and corresponding optimal action at each time step with respect to  confidence regions on the unknown parameter; 2) Thompson Sampling (TS) algorithm (a.k.a., posterior sampling),  \cite{thompson1933likelihood, kaufmann2012thompson,russo} which   samples an environment from the prior at each time step and selects the optimal action with respect to the sampled parameter. 
For the stochastic Linear bandit (LB) problems, there exist two well-known algorithms for LB are: OFUL or Linear UCB (LinUCB) and Linear Thompson Sampling (LinTS). \cite{Dani08stochasticlinear,rusmevichientong2010linearly,abbasi2011improved,moradipari2020stage} provided a regret bound of order $\mathcal{O}(\sqrt{T} \log T)$ for OFUL algorithm and \cite{agrawal2013thompson, abeille2017linear, moradipari2021safe,moradipari2020linear} provided a regret bound of order $\mathcal{O}(\sqrt{T} \log^{3/2} T)$ for LinTS in a frequentist setting.\\
\textbf{Multi-task Leaning and Meta-learning.~~~}
There has been an increasing attention on theoretically studying the  ability of a learner to transfer knowledge between different learning tasks, commonly referred to \textit{transfer learning} and applied to both the multi-task learning problem \cite{ando2005framework,pontil2013excess,maurer2013sparse,maurer2016benefit,cavallanti2010linear,du2020few} and the meta-learning problem \cite{baxter2000model,alquier2017regret,schaul2010metalearning,denevi2019learning,finn2019online,khodak2019adaptive,lee2019meta} in the past years. In particular, the goal of multi-task learning is to design an algorithm that performs well on a group of (possibly concurrent) tasks that share a similar representation (e.g., low-dimensional linear representation).   The goal of the meta-learning is to select an algorithm that can utilize a number of training tasks from a common environment in order to rapidly adapt to the new task that shares the same environment with the training tasks. We focus on the latter in this paper.  Recently, there have been a few works that study the multi-tasks learning in the bandit framework  \cite{calandriello2015sparse,azar2013sequential,zhang2017transfer,deshmukh2017multi,soare2014multi,liu2018transferable,yang2020provable,moradipari2021parameter}. In particular, the  recent works of \cite{hong2021hierarchical,kveton2021meta,hong2021thompson} study  meta-learning in multi-armed bandits problem with a Bayesian approach. In their settings, they consider a mixture Gaussian distribution as a prior distribution and propose a Thompson-Sampling algorithm with provable regret guarantees. However, in this work, we study a frequentist version of this problem and we propose a UCB-based algorithm. We also consider more general families of distribution for the mixture model.   


\section{Problem Formulation}

In this section, we briefly recall the preliminaries on stochastic linear bandit (LB) problem and previous results on meta-learning in LB, and then we present the multi-environment meta-learning setting considered in this work. 

\subsection{The Linear Stochastic Bandit (LB) Problem}\label{single-task-lb-problem}

In the LB problem, at each round $t \in [T]$, the learner is given an action set $\mathcal{D}_t \subseteq \mathbb{R}^d$ from which she chooses an action $x_t \in \mathcal{D}_t$ and observes a random reward \begin{align}
    y_t = x_t^\top \theta + \xi_t \label{reward_of_action}.
\end{align} 
In \eqref{reward_of_action}, the parameter vector $\theta \in \mathbb{R}^d$ is an \textit{unknown} but fixed reward parameter and $\xi_t$ is zero-mean additive noise. If provided with the knowledge of the true reward parameter $\theta$, the optimal policy at each round $t$ is to play the optimal action $x_t^{\star} = \arg\max_{x \in \mathcal{D}_t} x^\top \theta$ that maximizes the instantaneous reward. However, in the absence of such knowledge, the goal of the learner is to collect as much reward as possible, or minimise the  \textit{cumulative pseudo}-regret up to round $T$: 
\begin{align}
    R(T, \theta) = \sum_{t=1}^T x_\star^\top \theta - x_t^\top \theta. \label{cumulative_regret}
\end{align}
This classical setup defines {\it a single learning task} that takes $T$ rounds to complete.
Next, we will explore the setting where the learner is presented with {\it a sequence of learning tasks} in the form of LB problems that share probabilistic models for the unknown parameter vector $\theta$. By leveraging the structure shared between consecutive tasks, a so-called meta-learning algorithm introduces new inductive biases in the LB problem that allow the learner to transfer knowledge to  future tasks.

\subsection{Meta-Learning in LB } \label{sec:single-env-lb}

The problem of meta-learning for the linear bandit problem was first introduced by \cite{cella2020meta}. Their meta-learning setting consists of a sequence of  consecutive linear bandit problems that share the same {\it environment}, i.e., their parameter vectors $\theta_1,\dots,\theta_N,\dots$ are sampled independently from a task-distribution $\rho$ with a bounded support in $\mathbb{R}^d$ that is unknown to the learner.  The learner's goal is to leverage the task similarities (i.e., the fact that they share the same environment) in order to minimize the regret for a new task. In particular, \cite{cella2020meta} designed an algorithm that achieves a low regret on any new task after being trained over the data  provided by $N$ completed tasks. The goal is to control the so-called transfer regret incurred on the $(N+1)$-st task, defined as: \begin{align}
    \mathcal{R}(T)  = \mathbb{E}_{\theta \sim \rho} \left[ \mathbb{E}\left[R(T,\theta)\right]  \right]. \label{def:transferregetioflazaric}
\end{align}
Next we present the setting that we consider in this work, which is an extension of this setting to include multiple environments.

\subsection{Multi-Environment Meta-Learning in LB}\label{sec:multi-env-lb}

In the multi-environment setting, we assume that the consecutive tasks $i=1,\ldots,N$  originate from one of $m$ environments $\nu = 1,2,\ldots,m$ following a known multinomial distribution with probabilities $(p_1,p_2,\ldots,p_m)$. Conditioned on the environment $\nu$, the   task distribution for the parameter vector $\theta$ is denoted as $\rho_\nu$. The distributions $\rho_\nu$ have bounded supports in $\mathbb{R}^d$ and are not known to the learner. Instead of approaching each learning task independently, the learner collects information while interacting with the environments over $N$ consecutive tasks in order to perform meta-learning.
 Specifically, after completing the $i$-th task,  we store the whole interaction in a dataset $\mathcal{Z}_i = \{ (x_{i,t},y_{i,t}) \}_{t=1}^T$. Then, using the collected datasets from the first $N$ completed tasks, our goal is to design an algorithm that minimizes the regret for a new task with parameter $\theta_{N+1}$, without knowing  the environment $\nu$ from which $\theta_{N+1}$ is sampled. 
 In other words, we wish to design an algorithm that, after being trained over $N$ datasets, leverages  its past observations in order to introduce inductive biases to minimize the so-called \textit{transfer-regret} for task $N+1$: 
 
 \vspace{-0.4cm}
 \begin{small}
 \begin{align}
    \mathcal{R}(T) & = \mathbb{E}_{\nu } \left[\mathbb{E}_{\theta \sim \rho_\nu} \left[ \mathbb{E}\left[R(T,\theta)\right]  \right]\right] \label{transfer-regret}   \\& = \sum_{i=1}^m \mathbb{E}_{\theta \sim \rho_\nu} \left[ \mathbb{E}\left[R(T,\theta)\right] | \nu = i  \right] p_i.\label{chainrule_for_transferregret}
\end{align}
\end{small}
In \eqref{transfer-regret}, the outer expectation is with respect to the randomness over set of possible environments, the middle expectation is with respect to the the task parameters in each environment, and the inner expectation is with respect to the noisy components of the reward realizations. Note that due to this multi-environment setup, the knowledge gained from all the collected $N$ datasets $\mathcal{Z}_i, i=1,\dots,N$  may not transfer well to the new task parameter $\theta_{N+1}$ since the learner does not know the environment from which the new task originates. Accordingly, we need an algorithm that first decides to which environment the new task belongs. Then it uses an appropriate meta-learning scheme that considers the differences of the environments in order to leverage the task similarities to minimize the transfer regret. 
In order to provide training data for the algorithm to be able to distinguish between the environments (i.e., gain information regarding the unknown task distributions $\rho_\nu$), we  require that the learner is presented with a number of initial tasks with labeled environments in order to obtain a stationary behaviour in terms of  estimating a good bias parameter. Specifically, we introduce the following assumption. 
\begin{assumption}\label{ass:having_N_knowndataset}
We assume that for the first $N$ completed tasks, the learner has knowledge regarding the environment from which each task originates. In particular, we assume the learner has access to the sets $\mathcal{S}_\nu = \{ i : \theta_i \sim \rho_\nu, i = 1,\dots,N \}$ for $\nu = 1,\ldots,m$. We let $N_\nu = |\mathcal{S}_\nu|$.
\end{assumption}

\subsection{Model Assumptions}
Next, we present two more assumptions that are standard in the bandit literature \cite{abbasi2011improved}. 
\begin{assumption}\label{ass:sub-gaussian_noise}
For all $t$, $\xi_t$ are conditionally zero-mean R-sub-Gaussian noise variables, i.e., $\mathbb{E}[\xi_t | \mathcal{F}_{t-1}] = 0$, and $\mathbb{E}[e^{\lambda \xi_t} | \mathcal{F}_{t-1}] \leq \exp{(\frac{\lambda^2 R^2}{2})}, \forall \lambda \in \mathbb{R}$.
\end{assumption}

\begin{assumption}\label{ass:bounded_parameter}
There exists a positive constant $S$ and $L$ such that for every LB problem, $\norm{\theta}_2 \leq S$ and $\norm{x}_2 \leq L$ for every $x \in \cup_{s=1}^T \mathcal{D}_s$. Also, $x^\top \theta \in [-1,1]$, for all $x \in \mathcal{D}_t$. 
\end{assumption}
\vspace{-0.05cm}
\section{Background on Biased OFUL}\label{sec:backgroundkkonelsge}
\vspace{-0.1cm}
 
Before introducing our proposed MEML-OFUL algorithm for the setting introduced in Section \ref{sec:multi-env-lb}, in the following, we first review the OFUL algorithm and the biased version of OFUL, which our algorithm builds upon.

\vspace{-0.2cm}
\subsection{OFUL}

For the single  LB problem  in Section \ref{single-task-lb-problem}, we consider the OFUL algorithm  \cite{abbasi2011improved}. At each round $t \in [T]$, the algorithm uses the previous action-observation pairs and obtains a regularized least-square (RLS) estimate of $\theta$ as $\hat{\theta}_t = V_t^{-1} \sum_{s=1}^{t-1} y_s x_s $, where $V_t = \lambda I + \sum_{s=1}^{t-1} x_s x_s^{\top}$. Then, based on $\hat{\theta}_t$, OFUL builds a confidence set $\mathcal{C}_t(\delta) = \{ v \in \mathbb{R}^d : \norm{\hat{\theta}_t - v}_{V_t} \leq R \sqrt{d \log\left( \frac{1+tL^2/\lambda}{\delta} \right)} + \sqrt{\lambda}S := \beta_t(\delta)\}$ that includes a true reward parameter $\theta$ with probability at least $1-\delta$. Then, it plays an action $x_t$ by solving $x_t = \arg\max_{x\in \mathcal{D}_t} \max_{v\in \mathcal{C}_t} x^\top v$. For this algorithm, \cite{abbasi2011improved}  proves a high probability regret bound of order $\mathcal{O}(d \sqrt{T} \log(\frac{TL^2}{\delta}))$.


\subsection{Biased OFUL}
For a single  LB problem, \cite{cella2020meta} studies the biased version of the OFUL algorithm, called BIAS-OFUL. In particular, given a bias parameter $h \in \mathbb{R}^d$ for the true reward parameter $\theta$,  at each round $t \in [T]$,  the RLS-estimate $\hat{\theta}_t^h$ such that $\hat{\theta}_t^h = V_t^{-1} \sum_{s=1}^t x_s (y_s - x_s^\top h) + h$. Then, given an oracle that computes $\norm{h-\theta}_2$ for their algorithm, they show that they can build a confidence region $\mathcal{C}_t^h(\delta) = \{ v \in \mathbb{R}^d : \norm{\hat{\theta}_t^h - v}_{V_t} \leq R \sqrt{d \log\left( \frac{1+tL^2/\lambda}{\delta} \right)} + \sqrt{\lambda} \norm{h - \theta}_2 \}$ such that $\theta \in \mathcal{C}_t^h$ with probability at least $1-\delta$. They also adopt the same action selection rule as the one in OFUL, and using the Corollary 19.3 of \cite{lattimore2020bandit}, they provide an upper bound for the expected regret of their algorithm.
\begin{Proposition}[Lem. 1, \cite{cella2020meta}]
Under Assumptions \ref{ass:sub-gaussian_noise}, \ref{ass:bounded_parameter}, and considering $\lambda \geq 1$, the expected regret of the BIAS-OFUL is bounded as: 

\vspace{-0.5cm}
\begin{small}
\begin{align}
    \mathbb{E}&[R(T,\theta_\star)]  \leq   C \sqrt{T  d \log(1 + \frac{TL}{\lambda d})} \nonumber\\& \left(R \sqrt{d \log(T + T^2L/(\lambda d))} + \sqrt{\lambda} \norm{h -\theta_\star}_2  \right), \label{lazaric_expected_regret}
\end{align} 
\end{small}
where $C>0$ is a universal constant factor.
\end{Proposition}

It can be seen in \eqref{lazaric_expected_regret} that having a good bias parameter $h = \theta_\star$ brings a substantial benefit with respect to the regret (as $\lambda \rightarrow \infty$, the regret will tend to zero) in comparison to the unbiased case where $h = 0$.

Then, \cite{cella2020meta} adopts the BIAS-OFUL algorithm for the meta-learning setting described in Section \ref{sec:single-env-lb}. They adopt from \cite{denevi2018learning,denevi2019learning}, the idea of adding a bias parameter in a sequence  of the tasks that share the same environment and apply it to the linear stochastic bandit framework. Specifically. they show that for the meta-learning problem introduced in Section \ref{sec:single-env-lb}, running BIAS-OFUL with a bias parameter $h = \Bar{\theta} :=  \mathbb{E}_{ \theta \sim \rho}  [\theta]$ would significantly speed up the process of learning (i.e., lower regret) with respect to the unbiased case. This holds for a family of task-distributions where the second moment is much larger than the variance as formalized below.

\begin{assumption}\label{ass:task_variance_secondmomen}
The task-distribution $\rho$ satisfies:
\begin{align}
 \text{Var}_{\bar{\theta}} =  \mathbb{E}_{\theta \sim \rho} [\norm{\theta-\Bar{\theta}}_2^2] \ll \mathbb{E}_{\theta \sim \rho} [\norm{\theta}_2^2] =  \text{Var}_0, \nonumber
\end{align}
\end{assumption}
Overall, they show the following upper bound for the expected transfer regret defined in \eqref{def:transferregetioflazaric}. 

\begin{Proposition}[Lem.2, \cite{cella2020meta}] \label{props:bounding_lazaric_transfer_regret}
Let Assumptions \ref{ass:sub-gaussian_noise}, \ref{ass:bounded_parameter} hold, and fix $\lambda = \frac{1}{T \text{Var}_h}$. In the case where the tasks share the same environment $\rho$ satisfying Assumption \ref{ass:task_variance_secondmomen}, the expected transfer regret of BIAS-OFUL with a bias parameter $h$ is bounded as:

\begin{small}
\begin{align}
    \mathbb{E}_{\theta \sim \rho} &\left[ \mathbb{E}\left[R(T,\theta)\right]  \right] \, \leq \,
    d C\sqrt{T \log\big(1 + \frac{T^2 L (\mathbb{E}_{\theta \sim \rho} [\norm{\theta-h}_2^2])}{d} \big)} \nonumber
\end{align}
\end{small}
\end{Proposition}
 They also propose two strategies to estimate the bias parameter $h$ within the meta-learning setting in order to minimize the transfer regret. In particular, they show that if they can estimate the bias parameter $h$ equal to $\bar{\theta} $ with the meta-learning approach, then according to the Assumption \ref{ass:task_variance_secondmomen}, they substantially benefit from the task similarity compared to learning each task separately, i.e., compared to choosing $h=0$.


 \begin{algorithm}[th]
\caption{MEML-OFUL algorithm for task $N+1$}
  \label{alg:MEML-OFUL}
\textbf{Input:} $\lambda >1$, $T_0$, $T$,  datasets of $N$ completed tasks, \\
 Set $\hat{h}^1_{N+1}$ and $\hat{h}^2_{N+1}$ according to \eqref{seting_highvariance_bias_param}

\For{$t=1,\dots,T_0$}
{ 
Randomly choose $x_{N+1,t} \in \mathcal{D}_t$, and observe the reward $y_{N+1,t} = x_{N+1,t}^\top \theta_{N+1} + \xi_{N+1,t}$. \\
Compute $\hat{\theta}_{N+1,t}= V_{N+1,t}^{-1} \sum_{s=1}^{t-1} y_{N+1,s} x_{N+1,s}$ .
}

Select the bias $\hat{h}_{N+1} = \arg\min_{j \in \{1,2\}} \norm{\hat{\theta}_{N+1,T_0} - \hat{h}^j_{N+1} }_2^2$.

\For{$t=T_0,\dots,T+1$}
{
Build a confidence region $\mathcal{C}_t^h$ with a bias $\hat{h}^j_{N+1}$\\

Play $x_{N+1,t} = \arg\max_{x \in \mathcal{D}_t}\max_{v \in \mathcal{C}_t^h} x^\top v$\\

Observe reward $y_{N+1,t} = x_{N+1,t}^\top \theta_i + \xi_{N+1,t}$\\

Update
$\hat{\theta}_{N+1,t+1} = (\lambda I + V_{N+1,t+1})^{-1} \sum_{s=1}^{t} x_{N+1,s} (y_{N+1,s} - x_{N+1,s}^\top \hat{h}_{N+1}) + \hat{h}_{N+1}$\\

Update $V_{N+1,t+1} =  \sum_{s=1}^{t} x_{N+1,s} x_{N+1,s}^{\top}$
}
\end{algorithm}

 \section{Multi-Environment Meta-learning Algorithm (MEML-OFUL)}
\label{sec:algorithmdesctiopn}
MEML-OFUL builds on BIAS-OFUL to address the case where the learning tasks can  originate from multiple environments as explained in Section \ref{sec:multi-env-lb}. The summary of MEML-OFUL is presented in Algorithm \ref{alg:MEML-OFUL}.
In particular,  in order to minimize the transfer regret in \eqref{transfer-regret}, we employ the idea of applying a bias parameter for each task-distribution $\rho_\nu$ within the meta-learning setting. For brevity, we will state the results for the case where $m = 2$, i.e., there are only two environments.  
The extension to the case where $m>2$ is straightforward.

One of the main challenges of the multi-environment meta-learning problem is that when a new task $\theta_{N+1}$ is sampled, the learner does not know from which task distribution this task originated, and hence which  bias parameter to apply.
In particular, for a new task $\theta_{N+1}$, there exist two bias parameters $\hat{h}^1_{N}$ and $\hat{h}^2_{N}$ from  previously completed tasks in each environment, which can be used to transfer information to the new task. If the leaner selects the wrong bias parameter, then the regret of the new task could be larger than that of the unbiased case (i.e., independent learning of each task).
 To handle this issue, MEML-OFUL performs a pure exploration phase for the first $T_0$ rounds of a new task in order to calculate the RLS-estimate ${\hat{\theta}_{N+1, T_0}}$ of the new task parameter $\theta_{N+1}$. 
Then, it chooses the bias parameter that has the smallest square Euclidean distance from ${\hat{\theta}_{N+1, T_0}}$. It then runs BIAS-OFUL to complete the task and update the bias parameter.

We note that  the regret grows linearly with the length of the exploration phase.  However,  longer exploration  allows the learner to compute  more accurate RLS-estimates of the new task parameter, and hence minimize the misclassification probability (i.e., selecting the wrong bias parameter), presenting  a design trade-off. Note that even with a perfect estimate of $\theta_{N+1}$, misclassification can still happen as  $\theta_{N+1}$ might have non-zero mass in both distributions $\rho_1, \rho_2$.  As such, we need to carefully design the length of the pure exploration phase to be just long enough in order to compute a good  estimate  of the task parameter for classification, but not any longer so as to not adversely affect the regret.
After restricting the class of distribution $\rho_\nu$, our analysis in Theorem \ref{thm:upprbounding_transfer_regret_with_highvariance_appraoch} shows that we can set the length of the exploration phase such that it is constant with respect to $T$ and it inversely depends on the distance between the expected value ($\mu_i, i = 1,2$) of the task distributions (i.e., it captures the difference of the two environments).

We emphasize that in the case where the new task parameter $\theta_{N+1}$ has a non-zero probability of being sampled from both  $\rho_1$ and $\rho_2$, there  always exists a non-zero probability that MEML-OFUL chooses the wrong bias parameter, and hence suffers a larger regret. Therefore, in order to bound the regret,  we need to compute the probability that MEML-OFUL misclassifies the environment. 
To do so, we make the following assumption on the family of task-distributions we study.

 \begin{assumption}\label{ass:task-ditributions}
 We assume that for $i = 1,2$, the task-distribution $\rho_i$  is a multivariate  distributions on $\mathbb{R}^d$ such that any sample $x \sim \rho_i$ can be written as $x = \mu_i + z$, where $z \in \mathbb{R}^d$ has i.i.d. zero-mean, K-sub-Gaussian entries with a bounded support in $\mathbb{R}^d$, and $\mu_i \in \mathbb{R}^d$ is a fixed (but otherwise unknown) vector. We define the constant $\gamma = \norm{\mu_2-\mu_1}_2 $. Also, we assume $ \mathbb{E}_{\theta \sim \rho_i} [\norm{\theta - \mu_i}_2^2] \ll \gamma$ for $i = 1,2$.
 \end{assumption}

Then, we employ the biasing idea proposed in \cite{cella2020meta}, which is based on averaging the RLS-estimate of the task parameters of the first $N$ completed tasks with labeled environment (i.e., $\hat{\theta}_{j,T},  ~j=1,\dots,N$) without considering any bias . In particular, for a new task $N+1$, we set 

\vspace{-0.4cm}
\begin{small}
\begin{align}
    \hat{h}^1_{N+1} = \frac{1}{N_1} \sum_{j \in \mathcal{S}_1} \hat{\theta}_{j,T}, ~ ~    \hat{h}^2_{N+1} = \frac{1}{N_2} \sum_{j \in \mathcal{S}_2} \hat{\theta}_{j,T}. \label{seting_highvariance_bias_param}
\end{align}
\end{small}
\noindent Then, after the pure exploration phase, 
the algorithm decides to use $\hat{h}^1_{N+1}$ based on the Euclidean distance from the RLS-estimate of the new taks parameter (or similarly, $\hat{h}^2_{N+1}$). 
%

\begin{figure*}
     \centering
     \includegraphics[width=0.25\linewidth]{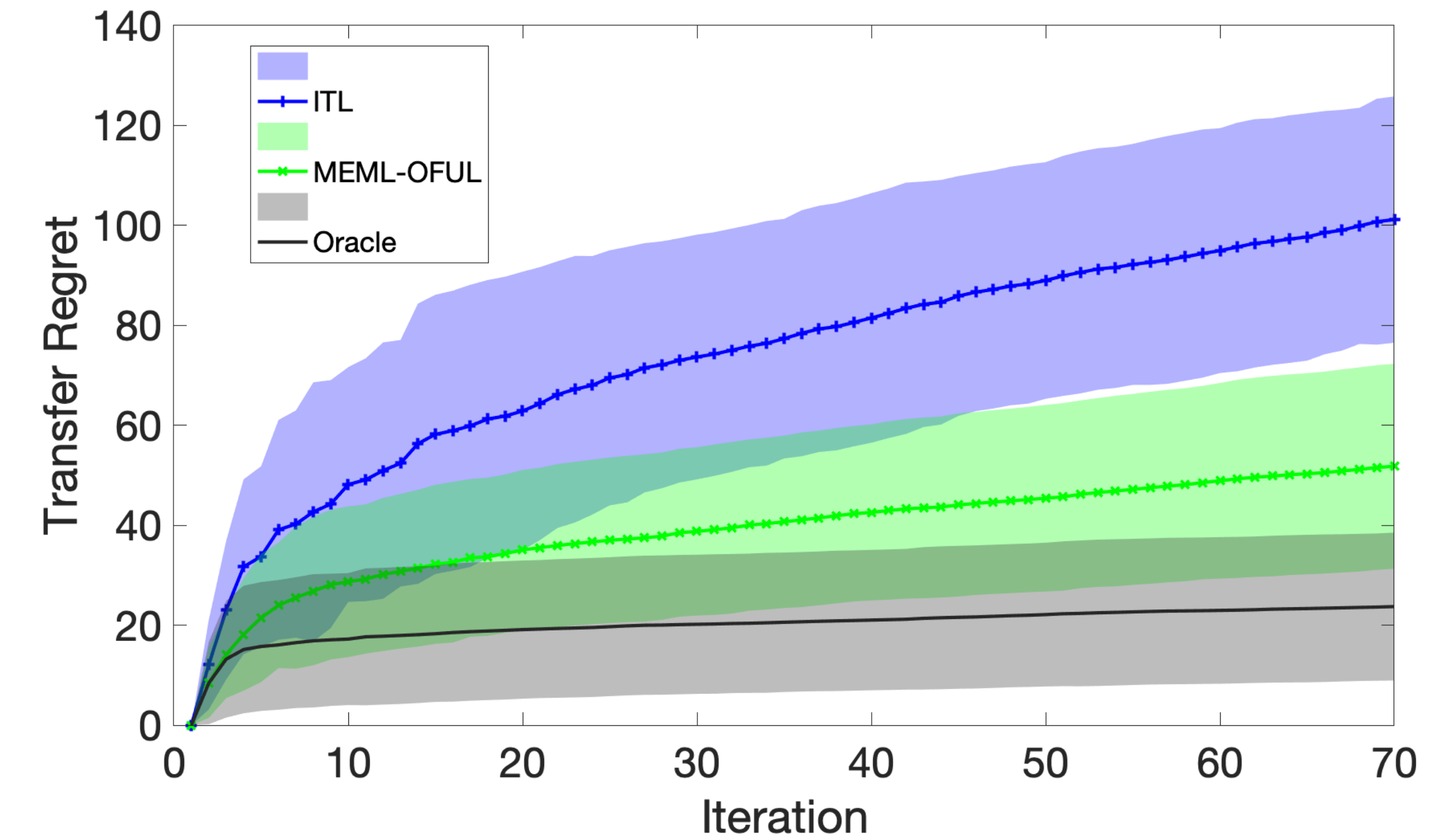}
      \includegraphics[width=0.25\linewidth]{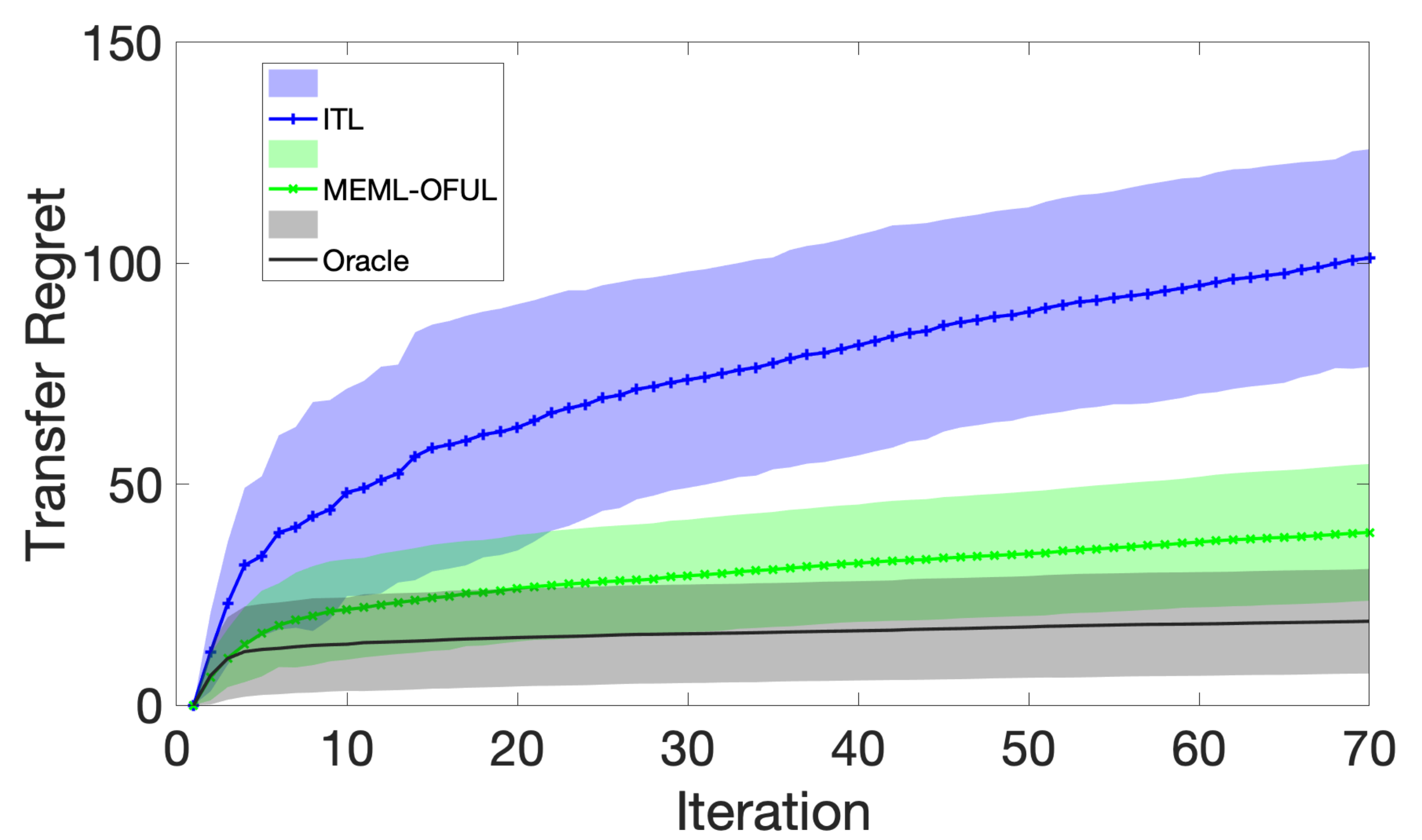}
          \includegraphics[width=0.25\linewidth]{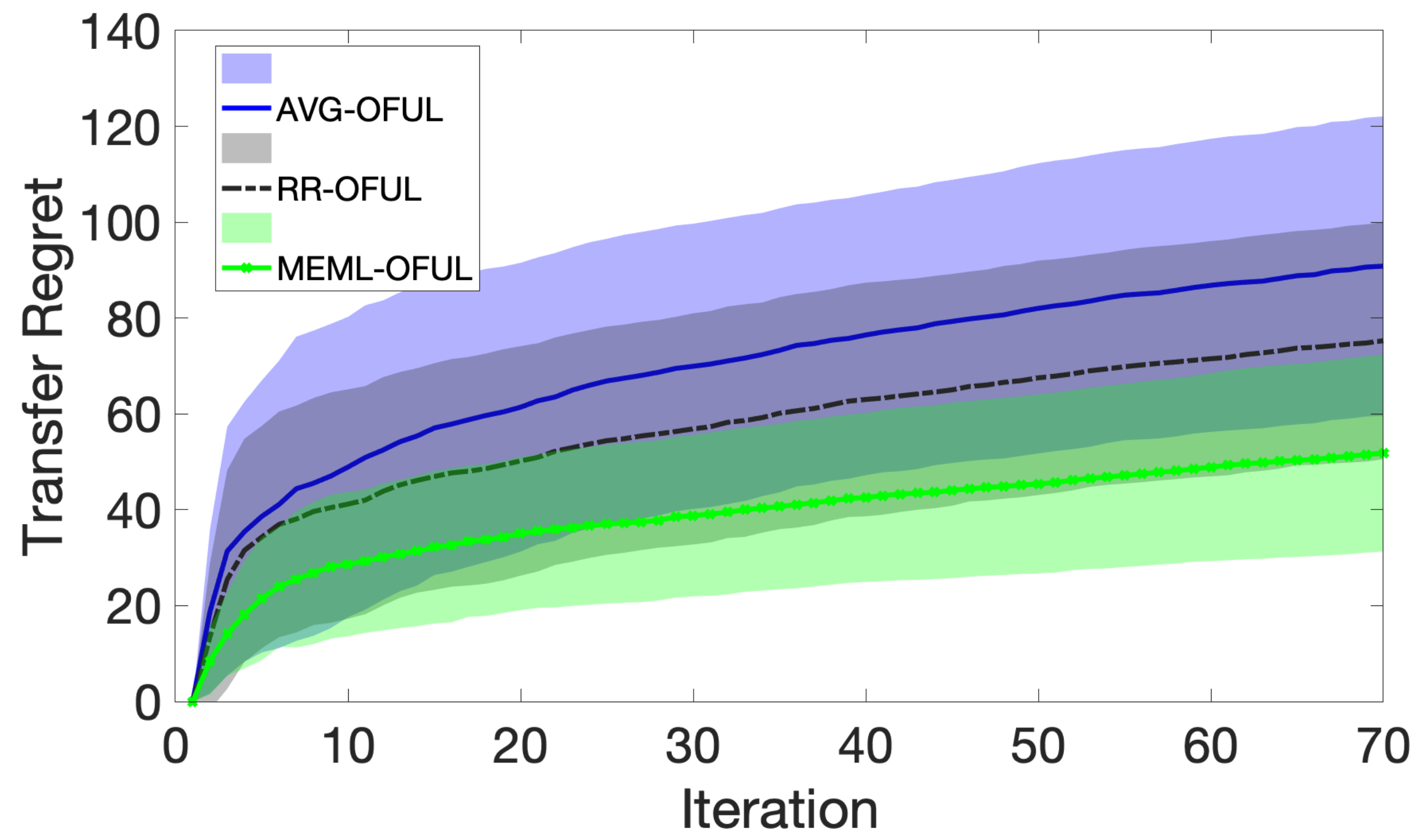}
         \caption{ Transfer regret of MEML-OFUL versus ITL and Oracle algorithm measured after $N_1=10$, $N_2=10$ training tasks. Left: $\lambda =1$; Middle: $\lambda =200$; Right: Comparing the transfer regret of our algorithm versus AVG-OFUL and RR-OFUL algorithms from \cite{cella2020meta} for the mixture distribution with $\lambda = 1$.}
         \label{fig:regret_bounds}
  \end{figure*}

\section{ Regret Bound } \label{metaleanring:highvariance}
 
In this section, we show the following bound on the transfer regret in \eqref{chainrule_for_transferregret} for MEML-OFUL algorithm.  

\begin{theorem}\label{thm:upprbounding_transfer_regret_with_highvariance_appraoch}
Let Assumptions {\red 1-5} hold, and let the bias parameters be defined as in \eqref{seting_highvariance_bias_param}.  Then, for   given prior probabilities $\mathbb{P}(\nu=1) = p_1$, $\mathbb{P}(\nu=2) = p_2$, the transfer regret defined in \eqref{chainrule_for_transferregret} is upper bounded as follows 

\vspace{-0.35cm}
\begin{small}
\begin{align}
& \mathcal{R}(T)  \leq \sum_{i=1}^2  p_i( 2T_0 + \nonumber\\&d C\sqrt{(T-T_0) \log\left(1 + \frac{(T-T_0)^2 L \left(  \text{Var}_{\mu_i}^i + \tau^i_{N} \right)}{d} \right)} ) \nonumber
\end{align} 
\end{small}
where  $T_0 > \mathcal{O}\left( \frac{R\sqrt{d \log(4/\delta)}}{\gamma - 
     \sqrt[4]{K^2 d(\frac{1}{N_1} + \frac{1}{N_2})}-2 K \sqrt{ {\log(4/\delta)} }} \right)$ for large enough $N_1, N_2$ such that ${ \gamma^2}{} \geq K \sqrt{ d(\frac{1}{N_1} + \frac{1}{N_2})}$, and $\text{Var}^i_{\mu_i} = \mathbb{E}_{\theta \sim \rho_i} [\norm{\theta - \mu_i}_2^2]$. Moreover, we have with probability at least $1-\delta$ for $\delta \in (\frac{1}{ 2 e^{\frac{\gamma^2}{4 K^2}}},1)$, for $i=1,2$:
     
     \vspace{-0.3cm}
     \begin{small}
     \begin{align}
    & \sqrt{\tau^i_{N}}  =  \sqrt{\norm{\mu_i - \hat{h}_{N}}} \nonumber \leq   \mathcal{O} \bigg( \frac{2 S \log(2/\delta) \sqrt{\mathbb{E}_{\theta \sim \rho_i} [\norm{\theta }_2^2]}}{N_i} \nonumber\\& + \max_{j \in \mathcal{S}_i} \{  \frac{\beta_{j,T}(1/T)}{\sqrt{\lambda + \lambda_{\min} (V_{j,T})}} \}+ \delta \left( \gamma  +  \frac{2S}{N}\right) \bigg). \label{boundoftheregretoffirstapproach}
\end{align}
\end{small}
\end{theorem}

The last term in RHS of \eqref{boundoftheregretoffirstapproach} comes from the misclassification of the environment for the new task. In particular, as $N$ grows to infinity, the RHS of \eqref{boundoftheregretoffirstapproach} is dominated by $\max_{j \in \mathcal{S}_i} \{  \frac{\beta_{j,T}(1/T)}{\sqrt{\lambda + \lambda_{\min} (V_{j,T})}} \}+ \delta  \gamma$, where the first term comes from the variance of the environment $\rho_i$ and the second term is caused by the inevitable effect of misclassification of the environment. 
We note that the lower bound for the number of rounds in the pure exploration phase is constant with respect to the time $T$,  and it  inversely depends on the probability that MEML-OFUL misclassifies the environment. We have also shown a lower bound on the number of tasks with a labeled environment that our algorithm requires in order to reach a stationary behaviour in estimating the bias.

\section{Discussion}
\subsection{When it is beneficial to apply MEML-OFUL?} \label{discuss:whypursarebetterthanlazaric}
Defining a mixture distribution $\rho = p_1 \rho_1 + p_2 \rho_2$ with known mixing probabilities such that $p_1+p_2 = 1$,  one may ask the question: what happens if we apply the algorithm in \cite{cella2020meta}  to our multi-environment meta-learning setting? We answer this question in several steps.

Consider the BIAS-OFUL algorithm  in  \cite{cella2020meta} applied to a new single-environment task-distribution $\rho = p_1 \rho_1 + p_2 \rho_2$ with  expected value  $\mu = p_1 \mu_1 + p_2 \mu_2$. In order to leverage the task similarities, BIAS-OFUL requires this mixture task-distribution $\rho$ to satisfy  Assumption \ref{ass:task_variance_secondmomen}, i.e., that  the task-distribution has a non-zero expected value. Therefore, for any family of environments such that $\mu = p_1 \mu_1 + p_2 \mu_2 = 0$ (e.g., $\mu_1= - \mu_2$ and $p_1 = p_2 = 1/2$),   BIAS-OFUL would not necessarily out-perform independent learning. However, MEML-OFUL  will not encounter this problem, since it interacts with each environment separately, and as long as $\mu_1, \mu_2 \neq 0$, it leverages  the task similarities of each environment to bring substantial benefit  with respect to the unbiased case.

Next consider  multi-environment  settings with $\mu = p_1 \mu_1 + p_2 \mu_2 \neq 0$. What do we gain  from  adopting the  MEML-OFUL algorithm over applying the BIAS-OFUL algorithm to the mixture distribution $\rho$? To start at a high level, we consider the case 
 that the meta-learning algorithm perfectly estimates the bias parameter to be equal to the expected value of the environment from which the task is sampled. In this case,
 Proposition \ref{props:bounding_lazaric_transfer_regret} shows the following bound on  the transfer-regret  of the mixture distribution for BIAS-OFUL:
 
 \vspace{-0.3cm}
 \begin{small}
 \begin{align}
 d C\sqrt{T \log\left(1 + \frac{T^2 L \left( p_1 \text{Var}^1_{\mu_1} + p_2 \text{Var}^2_{\mu_2}\right)}{d} \right)},  \label{lazarictransferregretcoparison}  
\end{align} 
 \end{small}
 For the same case, 
from the result of  Theorem \ref{thm:upprbounding_transfer_regret_with_highvariance_appraoch}, we obtain the following bound on the transfer-regret of the MEML-OFUL algorithm:

\vspace{-0.3cm}
\begin{small}
\begin{align}
& \mathcal{R}(T)  \leq  p_1 (2 T_0 + d C\sqrt{(T-T_0) \log\left(1 + \frac{T^2 L \text{Var}^1_{\mu_1}}{d} \right)} )\nonumber\\& ~~~+ p_2 ( 2 T_0 + d C\sqrt{(T-T_0) \log\left(1 + \frac{T^2 L \text{Var}^2_{\mu_2}}{d} \right)})   \label{ourmemltransferregretcomparison}.
\end{align} 
\end{small}
Now, we know from Theorem \ref{thm:upprbounding_transfer_regret_with_highvariance_appraoch} that $T_0$ is constant with respect to the time horizon $T$, and since the $\log$ and square root functions are strictly concave, we can conclude from Jensen's inequality that for a large enough $T$, the regret bound in \eqref{ourmemltransferregretcomparison} is less than the one in \eqref{lazarictransferregretcoparison}. This shows that it cannot hurt to adopt the MEML-OFUL over the BIAS-OFUL algorithm in the multi-environment settings \textit{if} the number of training tasks $N$ is sufficiently large to provide a close to exact estimate of expected value for each environment.  That being said, adopting MEML-OFUL has several extra requirements: 
1)  MEML-OFUL requires at least $N_1 + N_2$ training datasets of labeled environment such that $\gamma^2 \geq K\sqrt{d (\frac{1}{N_1} + \frac{1}{N_2})}$ as stated in Theorem \ref{thm:upprbounding_transfer_regret_with_highvariance_appraoch}. Of course, just meeting this minimum training requirement   does not guarantee that MEML-OFUL would outperform BIAS-OFUL;
2) MEML-OFUL requires a pure exploration phase to estimate the task environment, during which it incurs linear regret; 3) MEML-OFUL requires  knowledge of at least a lower bound on $\gamma = \norm{\mu_2-\mu_1}_2$, since the length of the pure exploration phase  depends on it.

\section{Numerical Results}
We investigate numerically the  effectiveness of MEML-OFUL for the meta-learning setting in Section \ref{sec:multi-env-lb}  on synthetic data. As mentioned in Section \ref{sec:backgroundkkonelsge}, in order to build the confidence regions, the algorithm requires the value $\norm{\theta_i - \hat{h}_{i}}_2$ which we upper bound similar to \cite{cella2020meta}. 
We first generate two environments  in agreement with Assumption \ref{ass:task_variance_secondmomen} such that $\rho_1$ and $\rho_2$ are Gaussian distributions with means $\mu_1 = [1;1]$ and $\mu_2 = [3;3]$ and $\text{Var}^i_{\mu_i} = 1, ~ i = 1,2$.  In all the implementations, we used $T = 70, \delta = 1/T, R = 0.1$. The transfer-regret figures are averaged over $N=10$ test tasks, where each environment $\rho_i$ was sampled with probability $p_i = 1/2$. For the decision set $\mathcal{D}$, we follow a similar approach to the one in \cite{cella2020meta}.   In Figure \ref{fig:regret_bounds} the shaded regions show standard deviation around the mean.   
We plot the MEML-OFUL algorithm as well as the independent task learning (ITL) policy, which completes each learning task separately.  Also, we plot the Oracle policy that knows the mean of each environment $\mu_1$ and $\mu_2$.  The transfer regret shown in Fig. \ref{fig:regret_bounds} (left) are for $\lambda =1$  and Fig. \ref{fig:regret_bounds} are for $\lambda = 200$.  Moreover, we adopt the two proposed algorithm AVG-OFUL and RR-OFUL proposed in \cite{cella2020meta} to our setting in the mixture environment $\rho = (\rho_1+\rho_2)/2$ which is a Gaussian distribution with mean $\mu= [2;2]$. Figure \ref{fig:regret_bounds} (right) shows that our proposed algorithms out-perform  both algorithms in \cite{cella2020meta}, which supports our discussion in Section \ref{discuss:whypursarebetterthanlazaric}.

\newpage
\bibliographystyle{IEEEtran}

\bibliography{bibfile}

\end{document}